# Pluggable Social Artificial Intelligence
# for Enabling Human-Agent Teaming

**J. van Diggelen (PhD), J. S. Barnhoorn (PhD), M. M. M. Peeters (PhD), W. van Staal (MSc),**
**M.L. Stolk (MSc), B. van der Vecht (PhD), J. van der Waa (MSc), J.M. Schraagen (Prof., PhD)**
TNO
PO Box 23
3769 ZG Soesterberg
THE NETHERLANDS

{jurriaan.vandiggelen, jonathan.barnhoorn, marieke.peeters, martin.stolk, bob.vandervecht,
jasper.vanderwaa, jan_maarten.schraagen}@tno.nl

## ABSTRACT

*As intelligent systems are increasingly capable of performing their tasks without the need for continuous human input, direction, or supervision, new human-machine interaction concepts are needed. A promising approach to this end is human-agent teaming, which envisions a novel interaction form where humans and machines behave as equal team partners. This paper presents an overview of the current state of the art in human-agent teaming, including the analysis of human-agent teams on five dimensions; a framework describing important teaming functionalities; a technical architecture, called SAIL, supporting social human-agent teaming through the modular implementation of the human-agent teaming functionalities; a technical implementation of the architecture; and a proof-of-concept prototype created with the framework and architecture. We conclude this paper with a reflection on where we stand and a glance into the future showing the way forward.*

## 1.0   INTRODUCTION

Recent developments in Artificial Intelligence (AI) technology, computational processing power, and the availability of data have given rise to increasingly intelligent systems, i.e. entities capable of engaging in dynamic and goal-directed interaction with their environment [48]. Intelligent systems are often described in terms of their behaviour and capabilities: intelligent systems can *sense* their environment, *reason* about their observations and goals in order to make *decisions*, and *act* upon their environment [62]. Due to their processing speed and their vast and almost infallible memory, intelligent systems outperform humans in handling large amounts of (heterogeneous) data, dealing with complex problems, and rapid decision-making [59].

  A more recent development is that, for certain tasks, intelligent systems are capable of operating at high performance levels for extended periods of time without the constant need of human support, guidance, or intervention [55][4][36][41]. Concerns about the proliferation of intelligent systems in defence, healthcare, aviation, and other high-risk domains - the military domain being one of the most prominent domains under debate [32][8] - has ignited heated debates around the globe [48], [49], [6], [14].

  Central to these debates about intelligent systems is the term 'meaningful human control' [19][1][3][32]: How can intelligent systems be developed in such a way that humans remain in control of the behaviour and effects of said systems? Ultimately, such control is necessary to allow for human responsibility and accountability for potential outcomes of the deployment of intelligent systems.

  One way of approaching the realization of meaningful human control is human-agent teaming (HAT). HAT aims to fully benefit from a system's autonomous capabilities while still maintaining meaningful human control. This is accomplished by endowing an intelligent system with a variety of team behaviours, making the system anticipatory, sensitive, and responsive to the needs, wishes, intentions, control, and/or influence of other team members. Examples are: pro-actively sharing bits of information to maintain a minimally required extent of shared situation awareness; being (re-)directable at a higher level of abstraction (e.g. strategic or tactical); anticipating (human) team members' actions; or foreseeing potential problems, sending human teammates timely warnings, and asking them for assistance. Most common architectures for autonomous systems (such as 4D/RCS [1]) ignore teaming functionalities and place the focus on task-oriented Artificial Intelligence (TAI), such as planning and sensing. The purpose of this paper is to describe a method for pluggable Social Artificial Intelligence (SAI) which allows developers to complement an existing autonomous (TAI) system with the capability to team up with humans.

The framework, called SAIL (Social Artificial Intelligence Layer) can be added to an autonomous system any time during commissioning or, at a later stage, while the system is in use. SAIL provides an infrastructure and a library of common HAT behaviours, promoting reusability, problem decomposition, and adaptability. This paper discusses the application of SAIL from functional analysis to functional design, to system architecture design to technical implementation.

We start our discussion in Section 2 on functional analysis by distinguishing five dimensions that can be used to characterize a Human-Agent Team, such as spatial dispersion of the team members, time criticality, and communication characteristics. Using three illustrative examples from the defence domain (mine hunting, aerial surveillance, and robot-assisted house search), we will demonstrate the application of the framework. These scenarios impose different requirements on a HAT and illustrate the scope of the problem space representing military HAT applications.

We have extracted a number of common high-level functions needed to enable team collaboration between humans and autonomous systems. These functions form the main HAT-functions provided by SAIL. For example, we identify a *proactive communication* function that decides whether a particular piece of information is relevant for a human given the current task context, user state, and system capabilities. Another common function within HAT is an *explainable AI* function that allows an autonomous system to explain why it has chosen a certain course of action and select useful explanations to offer to its human user to increase its predictability. Seven common HAT functions are described in Section 3.

The HAT functions are realized in the SAIL software architecture in a modular way using dedicated SAI-components which can be plugged into the autonomous system. Communication between these components is facilitated by our newly developed Human-Agent Teaming Communication Language (HATCL). This language provides the constructs needed for all components (i.e. TAI, SAI, and human) to coordinate actions among one another, ultimately leading to a coherent team consisting of humans and autonomous (TAI) systems glued together by SAI components. Of particular concern is to enable a mapping from the concepts in the internal control logic in the autonomous system to HATCL and back. This is realized in SAIL by so-called *semantic anchors*. A description of the SAIL architecture, HATCL and semantic anchors is provided in Section 4.

Using the SAIL framework, we have implemented a prototype application in which a swarm of military surveillance drones running in the Gazebo simulation environment can be controlled in a meaningful way using HAT techniques. The prototype is described in Section 5.

Section 6 presents a conclusion and future activities.

## 2.0 ANALYSING HAT SCENARIOS IN VARIOUS DIMENSIONS

Teamwork and collaboration requires interaction and tuning, especially in a HAT. What type of interaction, coordination, and alignment is needed, strongly depends on the type of HAT and the context in which it operates. Therefore, we have identified five important dimensions that describe and define a specific HAT, and that can be used as a guidance when determining specific requirements for a given HAT.

## 2.1 HAT dimensions

We describe a set of five dimensions that affect the requirements of a specific HAT: environment, mission/task, team organization, team dynamics, and communication (Figure 2-1 provides an overview). The dimensions can be used in multiple ways: to evaluate the generic applicability of HAT solutions for a wide variety of tasks, teams, contexts, and situations; to support rapid prototyping and analysis of specific HAT cases and identify challenges and requirements; or to compare multiple scenarios or cases to one another in terms of complexity and/or required SAI functionality.

### 2.1.1 Environment

The environment dimension describes which part of the environment is (a) dynamic, as opposed to static, and (b) predictable, and to what extent. This dimension affects, among others, HAT requirements aimed at the establishment of (shared) situation awareness (SA). Situation Awareness is "the perception of the elements in the environment within a volume of time and space, the comprehension of their meaning, and the projection of their status in the near future" [20]. In teams, SA needs to be distributed optimally across team members. Acquiring and maintaining shared SA becomes increasingly difficult as the environment becomes more dynamic and unpredictable [21]. Furthermore, complex environments also make it more challenging to determine when or in which manner the autonomous system may reach the boundaries of its capability envelope [57], posing additional challenges for the human to quickly switch to manual control if needed.

### 2.1.2 Mission / task

The mission / task dimension describes four factors. First of all, the (a) duration of the work cycles - which can be short, long, repeated, continuous, or team lifespan [54] - affects training and evaluation requirements. For instance, when teams collaborate for brief durations, extensive training and / or preparation may be needed because there is little time for tuning or adjustments during actual task performance. Secondly, the (b) interdependency of the team (e.g. [47]) can affect the need for interpredictability (also see Section 3.2) and shared mental models of team members' capabilities and status as members depend on each other's performance and actions. The other two factors of the mission / task dimension are (c) time criticality, affecting communication requirements, and (d) risk. From a social perspective, increased risk may require artificial teammates to be more aware of human team members' emotional status, which can potentially affect task performance.

### 2.1.3 Team organization

The team organization dimension describes (a) the team's physical proximity, (b) the number of team members, (c) the team's adaptability [47], and (d) their skill and authority differentiation (e.g., the extent to which the team members have different specialisms and ranks, [54]) and (e) network structure. Different organizational set-ups require different behaviour from teammates. For instance, we know from the human-human teaming literature that in distributed teams, team trust is an important mediator of success, and can be increased through effective knowledge sharing and exchange behaviours [24][58]. Proactive communication in HATs, a requirement we elaborate on later, contributes to this. The team adaptability factor is described as the ability to alter a course of action or team repertoire in response to changing conditions and is thus especially important in dynamic and / or unpredictable environments.

### 2.1.4 Team dynamics

The team dynamics dimension contains (a) the temporal scope - a team may be standing, ad-hoc, to be formed in the future or having ceased to exist [54]. The team's temporal scope may have implications for the extent to which teams are capable of forming longitudinal reciprocal and personalized team relationships. For example, in ad-hoc teams, the artificial teammate may have little or no time to develop a user model, and needs to be able to quickly identify roles, capabilities, and responsibilities. This is related to human-awareness, a requirement we elaborate on later in Section 3. The second aspect of the team dynamics dimension is (b) the team's current development phase - which cycles trough commissioning, preparation, action, and debriefing / learning. The development phase of the team, and especially the extent to which they actively cycle through these phases, may affect the extent to which teams engage in after action review and reflection to incrementally improve their processes and procedures. For instance, teams that cycle through preparation, action and learning phases may benefit from the ability to make work agreements among the team members, e.g. about constraints imposed on artificial agents, a functionality that we have implemented a solution for in SAIL (see Sections 3 and 4).

**2.1.5    Communication**

The communication dimension describes (a) the communication streams, which may vary between many-to-many, one-to-many, or one-to-one, (b) the information richness in communication [12], and (c) the quality of the infrastructure, which may vary in reliability, bandwidth, and range. During the design phase of a HAT, the communication dimension needs to be taken into account in order to optimize how, for instance, (shared) SA will be maintained [50]. For instance, when working in an unreliable network, relevant information needs to be pushed whenever possible, whereas with a reliable network, the artificial teammate may take other teammates' workload into account when timing communication. Furthermore, reduced communication abilities may increase the need to extensive training and clear work agreements so that teammates know what to expect from one another during periods of limited communication.

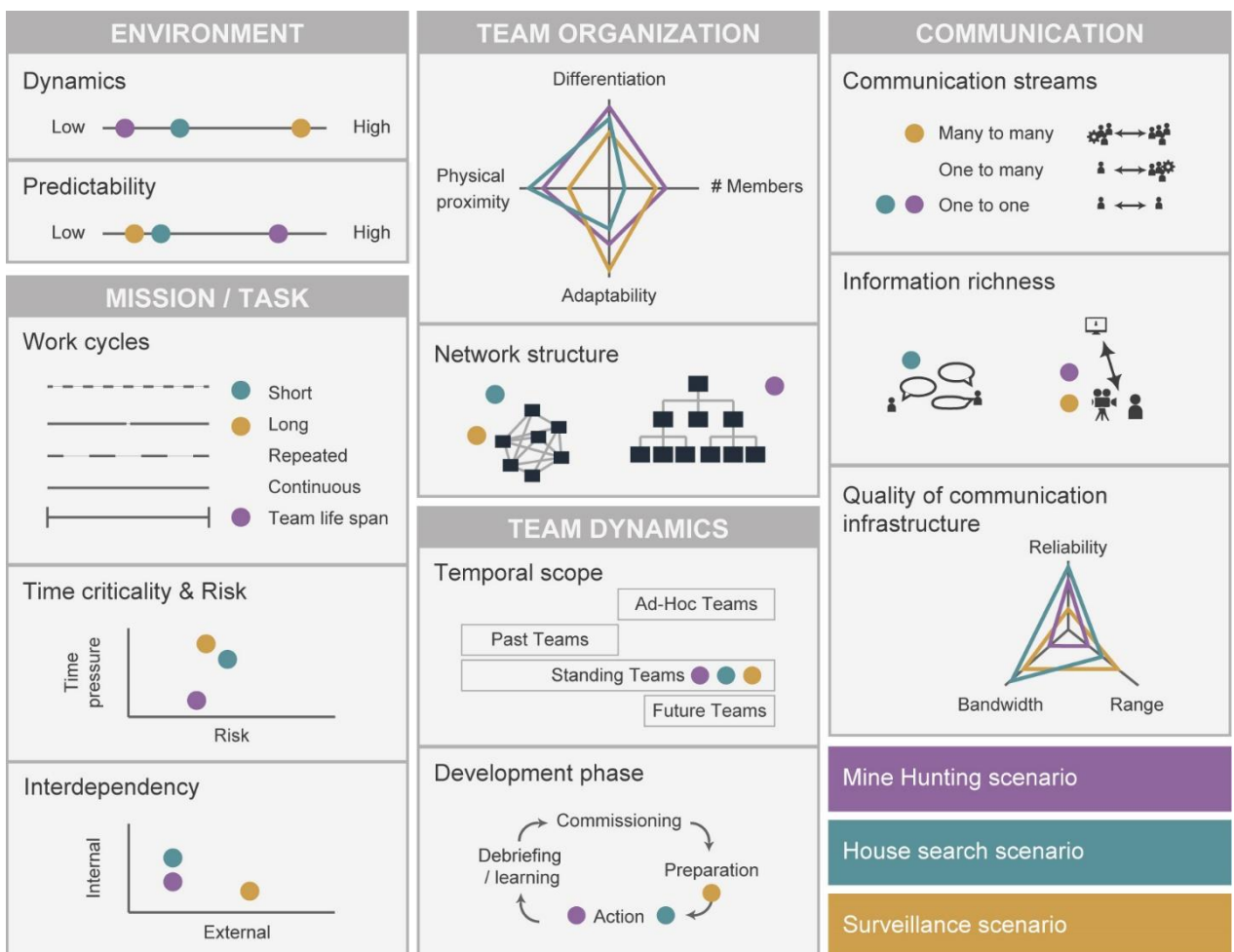

**Figure 2-1 Overview of the HAT dimensions, and the way each of the scenarios maps onto these factors. Scenarios are indicated by colour-coding (yellow, green, and purple).**

## 2.2    Example scenarios illustrating the dimensions

To illustrate the five dimensions presented in the above, we will discuss and analyse three typical defence scenarios. Figure 2-1 provides a visual overview comparing each of the scenarios with respect to the various team characteristics.

### 2.2.1 Surveillance scenario

In the Surveillance scenario, a swarm of unmanned aerial vehicles (UAVs) collaborates with a base commander to conduct a surveillance task around a small, temporary military base. The team has a low physical proximity, meaning that it requires the ability to efficiently exchange (shared) situation awareness. Differentiation of skill among the UAVs is low, allowing for quick back-up behaviour within the team. Furthermore, as the task concerns mainly surveillance, differentiation of authority between UAVs and the base commander is relatively low. In normal conditions, the UAVs can autonomously select sensors; plan and coordinate their flight paths; and choose what information to send to the base commander. However, when anomalies are detected or when the UAVs are unable to cope with a situation autonomously, the base commander quickly needs to (re)gain SA and decide on subsequent actions. This poses interesting requirements to the HAT: in normal conditions the marine officer has a low workload and SA, but in case of potential (often time critical) anomalies or unexpected situations he or she needs to be able to effectively transfer from 'out-of-the-loop' to 'in-the-loop' as quickly as possible (i.e. management by exception).

### 2.2.2 House search scenario

In the House search scenario, a single soldier teams up with a single unmanned ground vehicle (UGV) to perform a house search. Short work cycles are used as multiple houses may be searched subsequently. This means that there is room for (longitudinal) team development as the team repeatedly cycles from preparation to action and evaluation. Furthermore, the team has high internal, but usually low external dependency. The high internal dependency means that consistent behaviour, potentially enforced trough work agreements, and trust calibration are important. For instance, under what circumstances does the UGV perform below average in terms of situation assessment? And when trust is damaged, how can the artificial teammate repair it (e.g. [57][46][58])? The risk is high while time pressure is average, this means that it is probably most effective to proactively provide updates and request confirmation regularly while still considering the other teammate's current mental state. The house search team often needs to jump right into action and the environment is highly unpredictable, dynamic, and variable. This means that the team needs to be highly trained so that the team members work together seamlessly, even when they have little time to prepare for action while facing unpredictable situations.

### 2.2.3 Mine hunting scenario

In the Mine hunting scenario, personnel on a naval mine hunter teams up with a swarm of unmanned underwater vehicles (UUVs) to search for and dismantle sea mines. In this scenario, ample time is available for commissioning and preparation as there is usually relatively little time pressure. However, the availability of communication infrastructure during action is restricted due to limitations of underwater communication. This means that common ground is essential and work agreements are required that take unexpected situations into account, since agents may need to solve problems without being able to communicate. The mine hunting case is a typical example of a predictable, static environment with a team that is long-standing.

### 2.2.4 Applicability to other cases

The dimensions discussed above can also be used to describe and analyse other examples of military HAT. For instance, the goal keeper used to defend navy ships against incoming missiles is characterized by repeated work cycles with extreme time pressure and high risk. The wide range of potential HAT applications, situations, and settings makes it clear that HAT functions and solutions that aim to be generic need to be flexible, adaptable, and expandable. Together, SAIL and HATCL provide a platform to facilitate this. For the remainder of the paper, the surveillance scenario will be used as an illustrative running example.

### 2.2.5 Added value of HAT analysis

The HAT analysis can be used in many ways: to evaluate whether concept HAT solutions are as generic as imagined; to analyse specific HAT situations or cases and identify which capabilities are required for the artificial teammate; or to compare a set of scenarios or cases. In our experience, analysing cases, scenarios, and HAT solutions in this way can be very helpful in designing better human-agent interaction. Furthermore, it helped identifying the most common HAT functions, which we discuss in the next section.

## 3.0   COMMON HAT FUNCTIONS

This section describes the most common functions of agents participating in Human-Agent Teams. Our starting point are the three basic requirements for effective coordination [22][36]:

- *Common ground*: Team members must have shared beliefs about the world state, the goals, the conventions associated with the task at hand, etc.
- *Interpredictability*: team members must be capable of predicting each other's actions with a reasonable degree of accuracy.
- *Directability*: Team members must be capable of redirecting each other's behaviour.

By creating functionalities and behaviour capabilities for artificial team members that support each of these requirements, artificial agents can contribute to team processes in service of optimizing team performance. The following subsections offer insight in the way this can be achieved.

### 3.1   Common ground

In order for communication to be accurate and effective, participants in a conversation must establish proper common ground. Common ground refers to information which is mutually believed by all parties involved in a conversation [27]. Conversational efficiency is greatly enhanced through common ground as speakers need not explain all assumptions underlying their statements. Yet conversely, conversational effectiveness is greatly limited when speakers assume common ground, where it is lacking. Relying as a speaker on common ground with the audience is an invitation to cooperate [27] as the speaker appeals to the audience "to base their inferences not on just any knowledge or beliefs they may have, but only on the mutual knowledge or beliefs [shared by the conversational participants]" [10].

According to Stalnaker (2002), common ground relies on the speaker's presuppositions about the audience's common beliefs [53]. The common beliefs of the parties to a conversation are the beliefs they share, and that they recognize they share. By presupposing certain beliefs, the speaker takes the audience's understanding of such beliefs for granted as background information. By identifying the common ground of a conversation with the common belief of the participants, the presuppositions of an individual speaker can be identified with what the speaker believes to be common belief.

For HATs to establish and maintain common ground and thereby speed up the communication process, the following functionalities can be introduced in artificial team members.

#### 3.1.1   Shared situation awareness

An important challenge in developing HATs is to endow artificial team members with functionalities allowing them to effectively and efficiently share situation awareness (SA) within the team [19]. SA primarily refers to knowledge about the current state of the task environment, as well as team activities, team performance, and overall progression with respect to the team task. Such knowledge facilitates coordination and reallocation of tasks within the team, but can also be used for effective and efficient communication among the team members. SA can be used to explain dynamic goal selection, attention to appropriate critical cues, and future state predictions, but this requires for the parties involved to have access to the information needed to assess –and, through that, become aware of– the situation. Establishing shared situation awareness, then, means for team members to reason about the necessity of sharing certain information with the other members in the team, to develop a shared team "theory of the situation" [8]. To realize this, team members should decide to share relevant information, while withholding irrelevant information, so as to prevent their team members from becoming overloaded with information [46]. Deciding on the necessity of information for the other team members, requires an additional type of awareness, dubbed "intention awareness" by Howard and Cambria (2013), by which they mean "the process of integrating actors' intentions into a unified view of the surrounding environment" [31]. Intention awareness can also be used to reason about potential adversaries and other actors

### 3.1.2    Explanations

At times throughout a conversation, team members may fail to understand one another's actions or utterances. On such occasions, team members may require for additional explanations to gain understanding of the other team member's meaning, motivation, intention, or assumption. When creating artificial team members, developing functionalities enabling them to explain their behaviour, recommendation, or expression is often referred to as "explainable artificial intelligence (XAI)". XAI facilitates the disclosure of information by artificial team members to human team members. Compared to the sharing of situation awareness, XAI typically concerns more advanced functionalities, as it requires for the intelligent system to trace its internal line of reasoning, inference, classification, or input-output mapping, depending on the type of technology used. When done appropriately, XAI helps human team members understand the system's rationale underlying its behaviour and/or decisions. For example, an analyst who receives recommendations from a smart decision support system needs to understand why the algorithm has recommended a certain course of action.

As of recent, the research area of explainable AI (XAI) has exploded (e.g. [28]). In addition to the ability to offer a meaningful explanation for a specific human actor when needed, XAI also refers to the ability to ask for an interpretable explanation from a specific human actor when needed. So far, research centred primarily on the first type of ability. However, more integrative methods are under development, including bottom-up data-driven (perceptual) processes and top-down model-based (cognitive) processes [42]. Such methods could help assess the trustworthiness of the autonomous system's task performance and, subsequently, explain the foundation and reasons of this performance to establish trust calibration [58].

## 3.2    Interpredictability

Smooth collaboration can be enhanced by team members anticipating the interdependencies within the team. By predicting team members' task performance, team members can shorten waiting times, expedite task performance by aiding a team member in need of assistance, provide important just-in-time information to those who need it, and plan for contingencies if a team member may be unavailable or incapable of performing a task. To enhance interpredictability between team members, endowing an artificial team member with the following functionalities can be useful.

### 3.2.1    HAT training and longitudinal teaming

Team performance requires coordination between activities of each of the team members, under routine conditions as well as under novel conditions. Procedures, protocols, and doctrines are all artefacts created to foster team performance, as they increase interpredictability and support coordination. Procedural team training therefore focuses on the internalisation of procedures. However, procedures are often insufficient when the actual task deviates from the training task. On such occasions, teams have to improvise new ways of working together. One way of dealing with this challenge, is through team cross-training: team members switch roles with one another, so as to understand one another's roles and responsibilities [10]. Nikolaidis et al. (2015) investigated the use of cross-training in human-robot teams for assembly in the manufacturing process [43]. By iteratively switching roles between the robot and the human worker, the robot learned a model of human behaviour, describing the sequence of actions necessary for task completion and matching the preferences of the human worker.

Team training (both procedural and cross-training) aims to support the development of *shared mental models*: an overlapping understanding of one another's objectives, roles, tasks, activities, whereabouts, team structure, and so on [63]. Shared mental models enable team members to reason not only about their own situation, but also about that of their team members in the pursuit of their joint goal. Shared mental models, in other words, enable team members to predict a team member's performance on a particular task, potential need for help or information, risks of failure within the team, etc.

HAT training supports the development of shared mental models, by learning about one another's capabilities and limitations through experience, and optimizing the team processes needed to mitigate risks and limitations within the team. By endowing artificial team members with the functionalities required to engage in HAT training, artificial team members can learn to optimize their behaviour with respect to a team, i.e. the constellation of specific team members and their particular dynamics. In other words, team training facilitates the development of increasingly accurate mental models of team members, with the objective of increasing the team's ability to flexibly coordinate their activities and increase team performance. This, however, requires for artificial team members to be able to learn from team interactions, updating their models to better interpret the tasks, behaviour, and corresponding needs of their team members. This is also referred to in the literature as "interactive shaping" [38][43]. For example, Nikolaidis et al. (2015) found that cross-training improved mental model similarity, as well as the human worker's perceived robot performance and trust in the robot.

### 3.2.2    Proactive communication

Teamwork often entails the processing, interpretation, and analysis of large amounts of information. Based on their roles and/or expertise, responsibility for handling certain information sources and types is distributed across the various team members. Oftentimes, the results of the information processed by one of the team members are relevant to the activities of another team member, requiring the team members to communicate with one another. Communication in teams often aims to contribute to one of the following: (1) problem-solving, (2) structuring and coordination, (3) socio-emotional alignment, or (4) proactive communication [33].

An important challenge in teams is how team members decide to *proactively communicate*: when should an actor communicate what with whom. Such decisions are often based on shared mental models [63]. Interpredictability (facilitated by mental models) enables an agent to infer that its team member is currently working on a task that requires certain information, or that newly retrieved information affects the decisions and tasks of a team member. Based on such reasoning, the agent may decide to proactively share its knowledge with the respective team mate. Proactive communication entails team members providing one another with information on their own accord, i.e. without the need for a team member to explicitly request for that information to be shared. We distinguish between proactive communication to accommodate a team member's information need based on: (1) that team member's preferences (e.g. as learned from prior information requests), (2) knowledge about that team member's situation (i.e. based on a shared mental model), and (3) one's own potential need for assistance in the near future, requiring the envisioned assistant to be up to date with the situation at hand. There might be additional reasons and situations where proactive communication advances team performance, that we currently haven't thought of yet.

For an artificial agent to be able to reason about its *human* team members, requires for that agent to be *human-aware*. Human-awareness entails that intelligent actor(s) have access to information about human team members and their characteristics (e.g. preferences, tasks, capabilities and limitations, etc.). In addition to this information, the agent(s) also employ various functionalities aiding the maintenance of, reasoning about, and learning from such information. For example, functionalities related to human-aware computing include location monitoring, attention tracking, and trust calibration. Ultimately, human-awareness enables intelligent systems to predict the behaviour of human team members, fostering better shared mental models and interpredictability.

## 3.3    Directability

Directability entails the ability of team members to influence and/or control one another's behaviour, to accommodate adaptations in the team's activities, coordination, behaviour, and overall performance. Traditionally, in human-robot collaboration, directability entailed tasking of a robot by a human operator. However, as robots (and agents) become more capable of determining their own plans and activities to try and accomplish the team goals, human-robot collaboration gradually moves away from traditional tasking, and towards e.g. dynamic task allocation, shared initiative, and work agreements.

### 3.3.1    HAT communication

Being able to influence the behaviour of another actor first requires for the team members to be able to express themselves about more than simple information sharing, as was the case for common ground and interpredictability. We therefore need to extend communication to include speech acts that express the desire for another agent to perform a certain activity. In most agent communication languages, these speech acts are derived from "request" and "query" [23]. The request speech act asks another agent to perform a certain task or action, whereas the query speech act asks another agent to provide particular information. We will go deeper into the use of speech acts in the next section.

### 3.3.2 Work agreements

Work agreements are used to impose constraints on the autonomous behaviour of agents [16]. This forms one of the core building blocks of human-agent teams as they enable external directability on an agent's behaviour without compromising the agent's autonomy (also known as internal autonomy requirement [18][56]). Furthermore, they contribute to maintaining common ground and interpredictability, by enabling an explicit way to specify shared conventions on the agent's behaviour.

Work agreements are very similar to policies which have been applied as a teamwork coordination mechanism in [8]. We distinguish between two types of work agreements: obligations which describe which actions must be performed by an agent in a given context; prohibitions which describe which actions are not permitted to be performed by an agent in a given context. Examples of work agreements are

- UGV 2 has an obligation to notify the human worker when it detects a potentially hostile target
- UAV 3 has a prohibition to fly above the village

Work agreements can be applied for various purposes on different time scales. For example, one might specify a work agreement that specifies the current plan that is followed by the human and agent. On a longer timescale, a work agreement can be used to specify the Rules of Engagement which the system has to adhere to, or even military doctrine which lasts for the entire lifecycle of the system.

Work agreements can also be applied when designing the teamwork process itself by stating which communicative acts must be performed under which circumstances. For example, to specify the task division between the human and the agent, or to specify the level of human involvement in the process.

A more detailed (technical) description of work agreements is provided in the next section.

### 3.3.3 Dynamic task allocation and fit-for-purpose collaboration

As artificial team mates become more self-sufficient, it is at times unnecessary for a human to control, or even monitor, artificial agents at all times. The challenge is, however, that most artificial agents are capable of performing their task without the need for assistance or control under specific circumstances, whereas at other times, their performance degrades, or they malfunction altogether. As a result, the human-agent collaborative work relationship may vary across situations. To deal with this phenomenon, the team should be able to engage in dynamic task allocation and fit-for-purpose collaboration [25]. For example, the team should be able to shift between the following work relationships depending on the situation at hand:

- Parallel task performance: team members perform their tasks in coordination with their team members. They are capable of identifying, organising, and performing their own tasks and responsibilities without the need for assistance.
- Management by exception: team members can perform independently, yet when help or directions are required, they either ask for help from their supervisors or colleagues, or their team members notice a break-down and offer help proactively.
- Training / educating: team members are still in training and require constant feedback and monitoring so as to strengthen their understanding of the team task or goal, their team members, the roles within the team, and their own activities as part of the entire team performance.
- Tasking: team members are capable of performing a task once it is provided to them, but are incapable of determining their next task, as they are unaware of the encompassing team task or goal, their surroundings, role within the team, and/or team members.

## 4.0    SYSTEM ARCHITECTURE DESIGN FOR HUMAN-AGENT TEAMING

This section describes a functional architecture, called SAIL, in which the common HAT functions can be combined and configured to turn a set of autonomous agents and humans into a coherently working team. The next subsection describes the SAIL architecture

### 4.1    Social Artificial Intelligence Layer

SAIL (Social Artificial Intelligence Layer) is an environment in which HAT functionalities can be implemented in a modular way, i.e. using HAT modules. We distinguish between three types of components in a SAIL system:

- **Humans** in their ambient environment. For example, these may be professionals working in crisis management wearing mobile interaction devices such as smart watches and head-mounted displays supporting augmented reality; or these may be operators working in a control station with large information displays.
- **TAI** (task-oriented AI) components, i.e. technical AI components designed to optimally perform a certain task, but which may not be optimized for human interaction. These may be robots conducting surveillance in a certain area, cyber agents protecting vital ICT infrastructure against cyber threats, etc.
- **SAI** (Social AI) modules which serve as intelligent middleware aiming to transform task-oriented AI components and humans into a coherent human-agent team. Such components include machine learning technology that can decide how to exchange the right information at the right moment among the right actors [17], or AI message interpretation that can translate a high level command such as "secure the area" into commands that can be processed by the TAI component.

An overview of a SAIL configuration is depicted below.

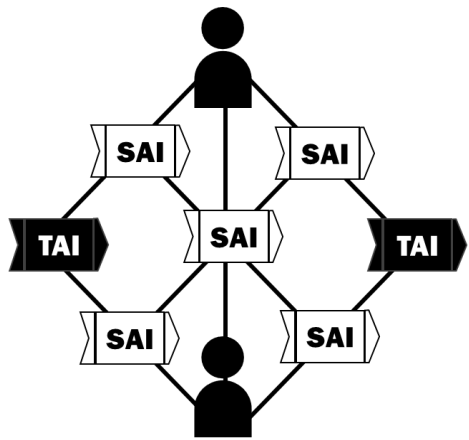

Figure 4-1 SAIL system architecture, where Task-oriented AI components (TAI) are mediated by Social AI Components (SAI) to allow interaction with humans.

Because the different components in SAIL can take on many forms depending on the requirements of the system, SAIL does not impose any constraints on their internal workings. The focus of SAIL therefore is on the interactions between the different components. These interactions are specified in a dedicated language called HATCL (Human-Agent Team Communication Language), which provides an abstract specification language of the possible information and control flows between the different components of a HAT. HATCL is abstract in the sense that it specifies the *information* in the message and its *illocutionary force* [50]. Illocutionary force refers to the intention of a speaker behind an utterance, e.g. obtaining information, directing, etc. HATCL messages are neutral with respect to interaction modality or graphical representation. A specification of the language is described in the next subsection.

## 4.2    HATCL

HATCL is inspired by FIPA-ACL, which has been developed as a language for communication between intelligent software agents [23]. To function as an effective means of communication between the different components in Figure 4-1, HATCL should satisfy a number of additional requirements which makes it suitable for Human-Agent Teaming.

**R1: HATCL should be capable of capturing the different abstraction levels at which machines and humans interpret and process information within a certain problem domain.**

HATCL should mediate between the different levels of abstraction of the various SAIL components. In the interaction between humans and SAI modules, a HATCL message should be capable of representing the information that stems from human input or that is outputted to humans. In other words, the language should (at least partly) be understandable by humans. This means that the language should align with human mental models [42]. One way to interpret this is to adopt Daniel Dennett's *intentional stance* [12]. This theory states that humans should comprehend the behaviour of artificial agents by attributing *goals*, *beliefs*, and *intentions* to them. The communication of such concepts should be facilitated by the HATCL language. In the interaction between TAI and SAI modules, the language should (at least partly) be understandable for machines. Therefore, the conceptualization should be alignable with the internal logic of the system. For example, when the system is based on internal logic that chooses its *actions* based on maximizing the expected *reward*, the language should be capable of communicating these concepts to be understandable for the system.

**R2: HATCL should allow for sending soft directives.**

To implement directability in a team of equal humans and agents, the rule *orders are orders* does not always apply. For example, consider the situation where the human commands an autonomous drone to fly back to the command post (CP). Various reasons exist in which the drone's responsive behaviour might deviate from simply following this command:

- **Incapability:** The drone might know that it is currently incapable of completing the action due to a reason unknown to the human (e.g. low battery level). In this case, we might want the drone to point out this problem to the human, rather than trying and failing.
- **Conflicting with team goal:** The drone might possess additional situation awareness that the given command would lead to an outcome which conflicts with the team goal. For example, that flying back to the CP would cause the drone to be subjected to hostile fire, which conflicts with the team goal to remain safe. If the drone can think of a better alternative, we might want the drone to propose this alternative instead.
- **Dealing with multiple orders:** In practical situations, the drone must deal with multiple orders which apply to different timescales and may even be in conflict with each other. For example, when the drone receives the order to fly to the CP, the drone may have to decide to finish its previous order first (e.g. to take a picture of an area of interest) or fly back immediately. Most likely, this depends on the time it takes to finish the previous order. In case of uncertainty, we might want the drone to discuss this problem with the human.

As the examples above illustrates, a directive used for communication between autonomous agents is fundamentally different than a directive used between objects in object-oriented programming language (e.g. remote method invocation). Jennings et al. have famously phrased this as: *Objects do it for free, agents do it for money* [36]. Nevertheless, defining the precise meaning of these soft directives remains a challenge, and is one of the major objectives of HATCL specifications.

**R3: HATCL should allow for specifying unambiguous work agreements.**

As argued in Section 3.3.1, work agreements form an essential part of HAT technology. Therefore, the specification of these work agreements in an unambiguous way is one of the main purposes of HATCL (also see Section 4.2.2).

**R4: HATCL should allow symbol grounding in various system architectures.**

The symbol grounding problem [15] refers to the problem of relating symbolic messages (such as HATCL messages) to internal structures that are processed by the agent (such as perceptions and plans). We use semantic anchors to formalize this relation. Different system architectures typically require different semantic anchors for the same HATCL message. For example, opaque deep learning networks require very different types of semantic anchors than rule-based systems.

To satisfy each of the requirements described above, we define the message structure (syntax) and its semantics (in terms of work agreements, ontologies, and semantic anchors).

### 4.2.1    The HATCL Message syntax

A HATCL message has the following structure: *<Performative, Sender, Receiver, In-reply-to, Content, Protocol, Ontology, Message-ID, Conversation-ID>*. Most of these fields, such as *Sender*, *Receiver*, *Message-ID*, contain meta-information used for routing the message. The fields *Performative*, *Content*, and *Ontology* are worthy of further explanation and are discussed below.

The *Performative* is used to denote the illocutionary force of a message, which could be:
- *Inform*: Provide another actor with information
- *Query*: Ask another actor for information
- *Subscribe*: Subscribe to information updates on a specific topic from another actor
- *Request*: Ask another actor to perform a certain task (acts as a single purpose work agreement)
- *Propose*: Propose a work agreement to another actor
- *Accept*: Accept the proposed work agreement
- *Reject*: Reject the proposed work agreement
- *Understood*: Acknowledge reception, and correct interpretation, of an inform message
- *Not understood*: Acknowledge reception, yet misinterpretation, of an inform message
- *Cancel*: Cancel a previously instantiated work agreement

This is the set of performatives currently included in HATCL. New performatives may be added as required by future applications. Each of these performatives has been defined both syntactically and semantically in our HATCL specification document.

The *content* of a message specifies what is actually communicated and can be specified in a query language, working agreement language, or assertion language. The content

```
{ "Performative" :      "Query" ,
  "Sender" :            "Hum1",
  "Receiver" :          "UGV1",
  "In-reply-to" :       "",
  "Content" :           "$.vehicles.*"
  "Protocol" :          "",
  "Ontology" :          "military_ont",
  "Message-ID" :        "msg13",
  "Conversation-ID :    "cnv-2" }
```

### 4.2.2    Work agreements

As work agreements between humans and autonomous systems impose well specified constraints on autonomous behaviour, they form the core building blocks of HAT technology; many message types in HATCL can be interpreted in terms of work agreements. For example, a HATCL message of the type *inform* is translated into the work agreement: *<Actor$_1$, Actor$_2$, upon receiving this message, O(send information of* *type x to Actor₁)>*, which states that by accepting this work agreement, Actor₂ commits to an obligation to immediately send information of type x to Actor₁.

Scientific research on the formalization of work agreements (also referred to as "social commitments") comes from the field of normative multi-agent systems [40][51]. A work agreement is an explicit agreement between two actors, specifying that one actor, denoted as *debtor*, owes it to another actor, denoted as *creditor*, to effectuate some *consequent* (e.g., refrain from or see to it that some action is performed or some objective is achieved) if the *antecedent* (e.g., some precondition) is valid [51]. Work agreements, in short, aim to specify permissions and obligations on agent behaviours. And so allow for the voluntary restriction of an actor's autonomy as proposed by another actor. Work agreements hold explicitly between two actors. Therefore, work agreements are sometimes compared to contracts. An example of a work agreement is: "Lawrence is obligated to notify Lisa about his change in intent, if he decides to pause his current task to switch to a more pressing task encountered along the way".

A work agreement is first and foremost a *voluntary* restriction on an actor's autonomy (also see Figure 4-2), as the actor receiving the *proposed* work agreement is also allowed to reject the work agreement. The *acceptance* of the work agreement, and hence the restriction on its autonomy, is completely voluntary. As soon as the debtor has accepted the work agreement, though, the debtor's autonomy is *conditionally* restricted: the debtor must satisfy the work agreement once the *antecedent* becomes valid. If the debtor fails to provide the consequent of an activated agreement before the deadline, this implies that the agreement has been violated. If the debtor succeeds to do so, the work agreement is satisfied. Furthermore, commitments can, in general, be cancelled by the actors involved (although this may be subdue to overarching rules).

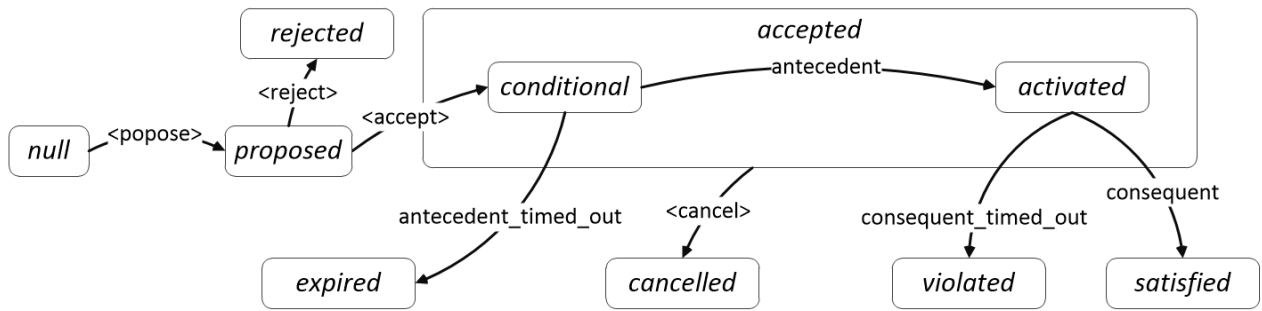

**Figure 4-2: The lifecycle of a work agreement as affected by the dynamics between the actors and events taking place in the environment**

A similar approach to work agreements are *policies*: "enforceable, well-specified constraints on the performance of a machine-executable action by a subject in a given situation" [16]. Policies come in two flavours: authorization policies, stating what is permitted, and obligation policies, stating what is obligatory in a given situation [16]. Policies do not hold between two actors, but instead are generally applicable to a set of actors. Therefore, policies are sometimes compared to laws. An example of a policy is: "Team members are obligated to notify their team leader about a change in intent if they decide to switch tasks after encountering a more pressing task".

The foundation of both policies and work agreements are normative rules in the form of deontic logic [59]. Deontic logic is used to reason about obligations and permissions. Therefore, deontic logic still forms the core of work agreements and policies, as it enables reasoning and verification.

### 4.2.3 Ontologies

Ontologies offer explicit, structured, and semantically rich representations of declarative knowledge. They consist of concepts (`classes'), and relations between them, to describe certain parts of the world [45]. HATCL uses a domain-independent "top ontology" and a domain-specific ontology to enable actors in the HAT to parse the messages they receive. Figure 4-3 shows the top ontology, consisting of relatively generic concepts, such as Actor, Plan, Goal, and Action. The domain-specific lower-level ontology would provide specific instances of tasks, actors, and plans particular to that domain.

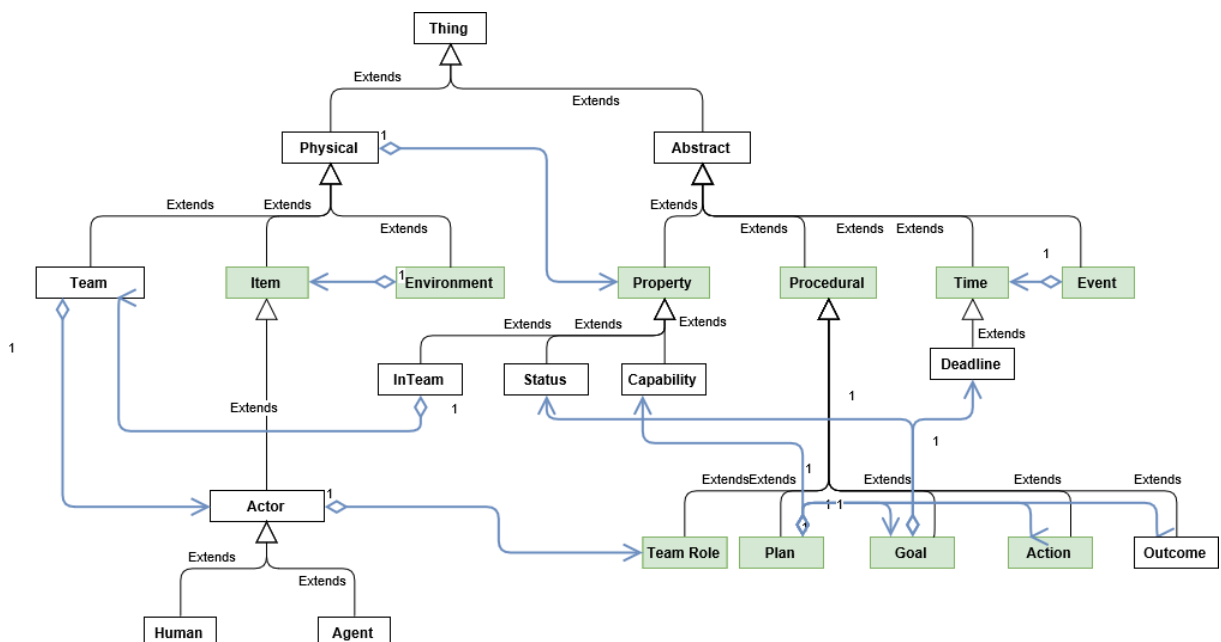

**Figure 4-3: Top-level domain-independent ontology used to specify work agreements and HATCL messages, facilitating coordination and communication between actors in the HAT**

#### 4.2.4    Semantic Anchors

Semantic anchors relate concepts and variables in HATCL to concepts and variables in the autonomous system. This translation operation is implemented inside the autonomous system. To illustrate this important principle in human machine teaming, we will start with a simple example.

**Example 1:** Suppose we have a human team-member controlling an autonomous system which runs the following code:

```
Repeat
        turn_left,
        turn_right,
        move_straight
Until false
```

Suppose the human teammember expresses the desire to prohibit left turns via work agreements. Right now, the autonomous system code is not suitable for that. Therefore, the developer makes the following code-update:

```
Repeat
        If tl_permitted then turn_left,
        turn_right,
        move_straight
Until false
```

Furthermore, a semantic anchor is created that maps the WA (specified in HATCL) to code that can be interpreted by the autonomous system:

```
        prohibited(turn_left) -> tl_permitted=false
```

**Example 2:** Assume now that the autonomous system's actions do not have the attribute `tl_permitted`. Instead, the system has a variable map indicating the desirability of each position on the ground, which is used to determine where to move. In this case, the anchor would access the variable map and set a low desirability on the coordinates on the system's left side.

As this example shows, multiple ways exist in which a semantic anchor can be realized. In general, the human interfaces work at a higher abstraction level than the internal control logic of the autonomous system. This means that semantic anchors whose information flows from the teaming software towards the system perform a translation into a lower abstraction level. Whereas the translation into a higher abstraction level occurs when the information flow is reversed. Note that this does not need to be the case nor are semantic anchors limited to a one direction flow of information. See the figure below for two examples of the functionality of semantic anchors. The top anchor simply acts as a gateway between an abstract variable in the teaming software and the system, as the abstract meaning is appropriate in both (see the example 1). Whereas the bottom anchor performs an actual translate operation and whose information flow is bidirectional (see example 2). Also note the presence of the API as a distinct entity.

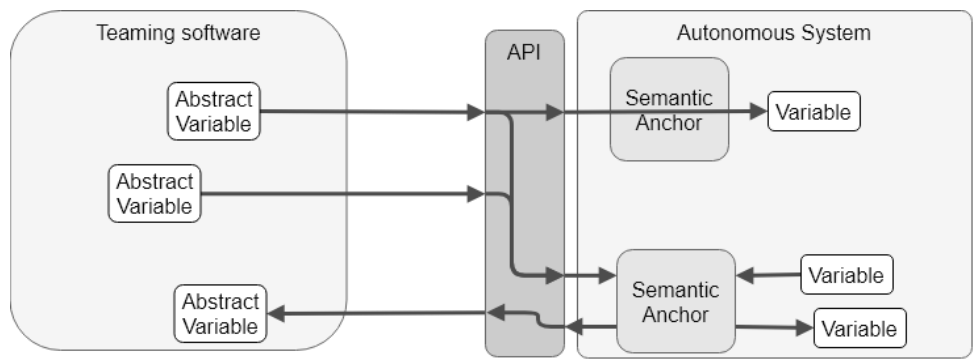

**Figure 4-4: Semantic anchors describe how abstract variables in the HATCL ontology are grounded in the variables available in the autonomous system**

Note that semantic anchoring can become very difficult or even impossible depending on the representations used. HATCL is based on the symbolic representation paradigm where each element corresponds to one entity. Neural Networks, which are widely used in AI applications such as autonomous driving and image classification, are based on distributed representations [29]. In these representations, each entity is represented by a pattern of activity distributed over many elements. How to map HATCL to distributed representations in a neural net remains an open question.

### 4.2.5 Software implementation of SAIL

SAIL is implemented using an open source distributed application framework, called Akka, which is used as a software wrapper around the various pieces of code, making it a coherent human-agent system. The SAIL components can be programmed in any language, and may run on any type of hardware (e.g. robots, head-mounted displays, mobile devices, sensors).

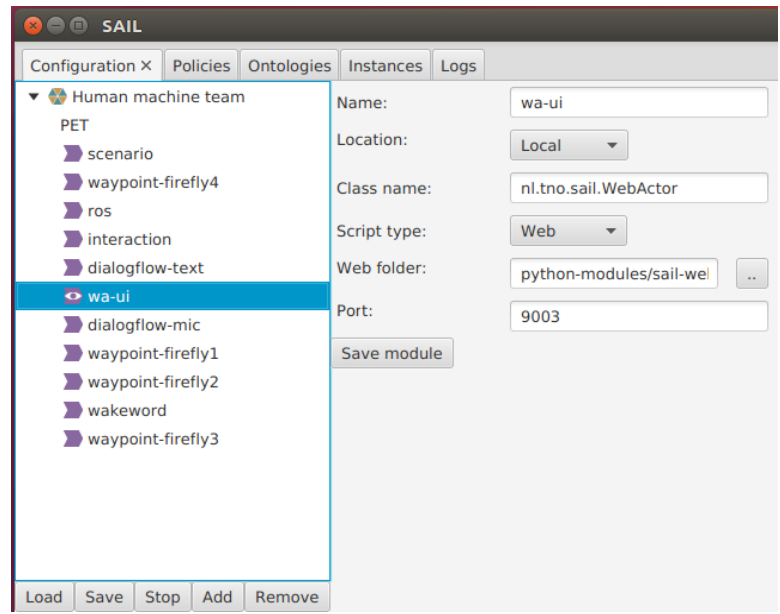

**Figure 4-5 SAIL development environment allowing the HAT engineer to add extra SAI modules and configure the system's set of fixed working agreements and ontologies.**

The figure above shows the Sail development environment. The different SAIL components can be added to the tree on the left-hand side of the window. The configuration of these components is done on the right-hand side. Other tabs can be used to specify the policies (working agreements), and ontologies that are shared within this system.

## 5.0 PROTOTYPE OF A HAT APPLICATION WITHIN SAIL

To demonstrate the application of the techniques discussed in the previous section, we used a case for aerial surveillance of a compound as described in Section 2.2.1. The proof of concept implements an initial subset of the proposed common HAT functions and shows that the underlying concepts and SAIL architecture translate into a viable HAT set-up. The surveillance scenario is recreated as a virtual environment. This simulation (implemented in Gazebo [35]) includes a 3D modelled military compound, a variety of potential threats in the vicinity of this compound as well as a swarm of UAVs that survey the surrounding area.

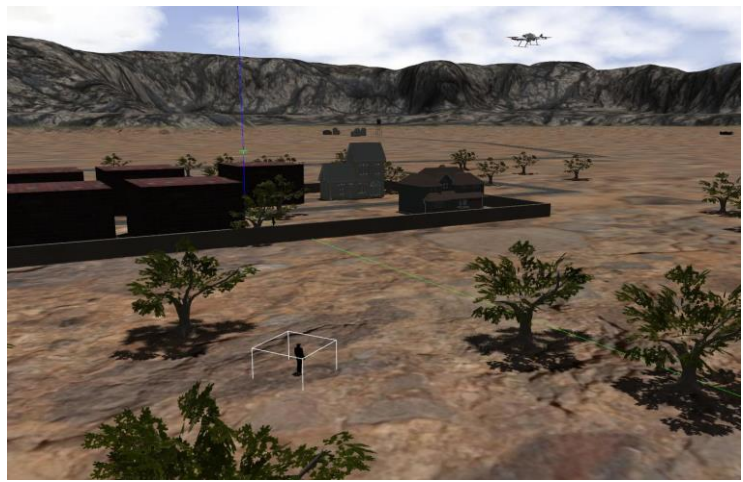

**Figure 5-1 the Gazebo simulation environment, which provides the backbone of the prototype but which is typically not visible as such to the base commander.**

Task-related functionality of the UAVs is implemented in TAI modules that cover capabilities such as waypoint navigation, path planning, object detection, video streaming, and threat identification. Using this set of TAI modules, each UAV is capable of autonomously scanning the surroundings of the military compound for suspicious activities.

To turn the UAV's into teammates, we added a Social Artificial Intelligence Layer (SAIL). This layer includes a set of SAI modules, including a multi-modal user interface, and offers the infrastructure to provide the middleware between the task-oriented modules and the human team members. In the proof of concept, we have implemented and combined a number of core HAT functions (as discussed in Section 3), namely those for shared situational awareness, proactive communication (ProCom), human-aware computing, HAT communication, and the work agreement mechanism.

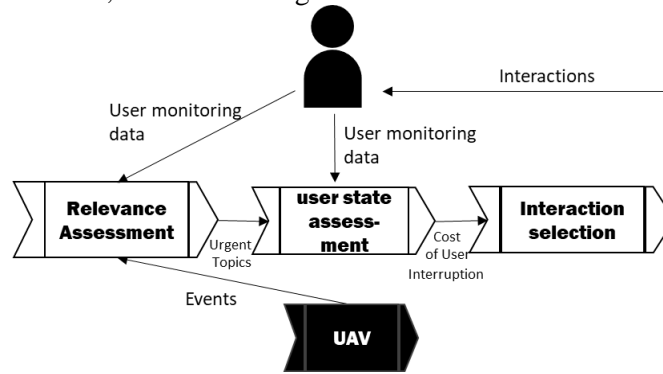

**Figure 5-2 An overview of the main SAI modules within the base protection HAT.**

The figure above shows the SAI modules that are related to proactive communication (ProCom). ProCom aims to establish a balance between the value of sharing information and the costs of imposing cognitive workload on the human team member. The autonomous system (the UAV) performs the task of aerial surveillance. Using a semantic anchor, events are published using HATCL and made available to the SAI module relevance assessment. This module regards each event as a topic and assesses its relevance to the human team member. For example, topics about the detection of a potential hostile contact are more relevant than topics about a friendly civilian contact. Also, when the human team member has explicitly asked for a certain topic, this topic has high relevance. The SAI module "user state assessment" builds up a value that indicates the cognitive task load of the human team member (how busy he/she is) and the situational awareness (what the human team member currently does and does not know). Based on this information, the SAI module interaction selection decides whether that information should be communicated to the human team member or not and in which way (e.g. using a textual message or using a voice-message). More information on content-based modality selection can be found in earlier work [17].

The combined SAI functionality manifests itself to the human team member in the form of an avatar. The avatar is able to maintain a dialogue with the human team member and act as an intuitive interface between him/her and the SAI components of the autonomous agents (e.g. UAVs). In addition the avatar acts as an intelligent information retrieval system, capable of accessing various information resources within the available SAIL modules. The avatar responds to speech, typed chat messages, and touch input from the human team member. Its actions can vary from information retrieval, engaging conversational dialogues with the UAVs, establishing work agreements with these SAI-plugged agents and displaying (task-related) content tiles on one of the available computer screens.

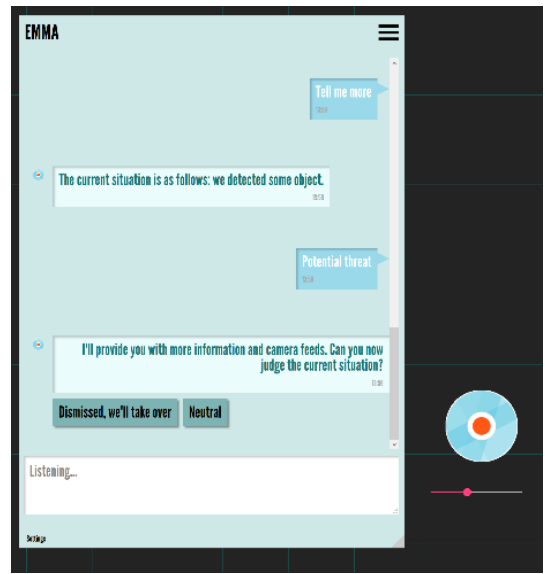

**Figure 5-3 The avatar representing the UAV swarm and the dialogue window enabling interaction.**

The figure above shows the avatar (the round circle on the right), and the dialogue window which interacts in a similar way as Google Allo (https://allo.google.com/). Natural language input can be inputted via speech or command buttons, and a history of the conversation is shown using text balloons. Besides replying to the human team member's input via text messages, the system can also open up additional windows to visualise information and provide new ways of interaction, such as maps, camera feeds of the robots, etc. Note that this can only be done if sufficient screen space is available.

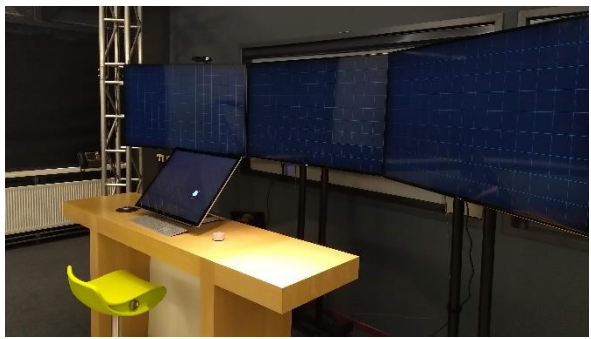
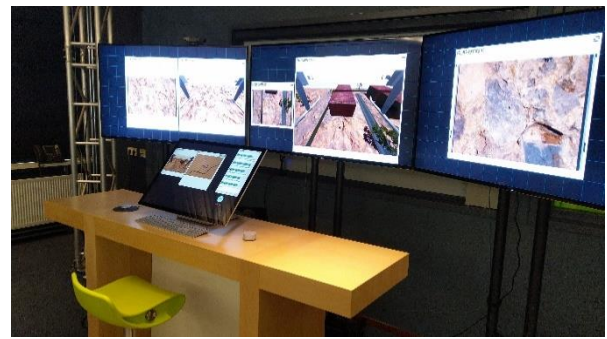

**Figure 5-4 The demonstration environment of the base protection HAT. When the autonomous system is not encountering problems, the base commander is not distracted with any information (as on the left hand side). When problems occur, the base commander can drill down on every piece of relevant task information.**

The figure above shows the set-up of the demo, consisting of three large information displays, a MS surface pro touch interaction device, microphone, and speakers. When the UAVs are functioning normally, the screens show as little information as possible, and only the avatar is visible (see the right hand side of the Figure). When an important topic arises, such as a suspicious contact, more and more information is exchanged between the UAV swarm and the base commander. The right hand side of the figure shows the situation in which the individual camera-feeds of the UAVs are shown to the base commander allowing the UAV and base commander to make a joint decision on how to classify the object. After the issue is solved, the topic becomes irrelevant again which implies that all camera-feeds close, and the interface returns to its calm state again.

The application described above illustrates a management-by-exception type of HAT. This means that the base commander is not bothered with superfluous information when this is not needed. However, upon request, or by system initiative, a rich interaction can be set up which drills down to the details of the matter at hand.

## 6.0 CONCLUSION

Human agent teaming is a problem with many faces. This paper is an attempt to combine a human factors, engineering, and military perspective on the issue. Our solution is based on the idea of a Social Artificial Intelligence Layer (SAIL), which is a framework for the development of HMT-concepts. The starting point of SAIL is that a HAT can be developed without changing the internal capabilities of the autonomous system. We have argued that these social capabilities are to some extent generic. Examples are functions for situation awareness, human awareness, explainable AI, working agreements and tasking. Within SAIL, HAT-modules are developed that construct these social capabilities. The modules are reusable in multiple domains.

We have demonstrated the use of SAIL by building an application for military compound protection using surveillance drones. The surveillance drones where simulated using the robot simulation environment Gazebo, and SAIL was used to build a teaming layer on top of it. The approach resulted in a system which embraces a management by exception type of HAT: no information reaches the human teammember, unless a problem arises which requires negotiation.

We believe that our approach is promising and an important step towards developing and prototyping human agent teaming applications in the defence domain. We identify three directions for future work. Firstly, we intend to develop more complex types of semantic anchors, and explore which autonomous system architectures enable which types of anchors. This allows us to discover the boundaries of HAT applications, as only those types of information can be communicated which can actually be anchored in the agent. Secondly, we intend to explore validation methods for HAT applications, which also take the long term aspects of teaming into account. Thirdly, we intend to explore combinations of HAT interaction with more immersive interaction techniques such as tele-presence.

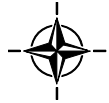